\documentclass{article}

\usepackage[english]{babel}

\usepackage[letterpaper,top=2cm,bottom=2cm,left=3cm,right=3cm,marginparwidth=1.75cm]{geometry}

\usepackage{amsmath}
\usepackage{graphicx}
\usepackage{listings}
\usepackage{subcaption}
\usepackage{amsfonts}

\usepackage{orcidlink}

\usepackage{multirow}
\usepackage{booktabs}

\title{Restrictive Hierarchical Semantic Segmentation for Stratified Tooth Layer Detection}
\author{Ryan Banks \orcidlink{0009-0008-6504-7084}, Camila Lindoni Azevedo \orcidlink{0000-0002-9142-257X}, Hongying Tang \orcidlink{0000-0003-2534-737X}, Yunpeng Li \orcidlink{0000-0003-4798-541X}}

\begin{document}
\maketitle

\begin{abstract}

\noindent Accurate understanding of anatomical structures is essential for reliably staging certain dental diseases. A way of introducing this within semantic segmentation models is by utilising hierarchy-aware methodologies. However, existing hierarchy-aware segmentation methods largely encode anatomical structure through the loss functions, providing weak and indirect supervision. We introduce a general framework that embeds an explicit anatomical hierarchy into semantic segmentation by coupling a recurrent, level-wise prediction scheme with restrictive output heads and top-down feature conditioning. At each depth of the class tree, the backbone is re-run on the original image concatenated with logits from the previous level. Child class features are conditioned using Feature-wise Linear Modulation of their parent class probabilities, to modulate child feature spaces for fine grained detection. A probabilistic composition rule enforces consistency between parent and descendant classes. Hierarchical loss combines per-level class weighted Dice and cross entropy loss and a consistency term loss, ensuring parent predictions are the sum of their children. We validate our approach on our proposed dataset, TL-pano, containing 194 panoramic radiographs with dense instance and semantic segmentation annotations, of tooth layers and alveolar bone. Utilising UNet and HRNet as donor models across a 5-fold cross validation scheme, the hierarchical variants consistently increase IoU, Dice, and recall, particularly for fine-grained anatomies, and produce more anatomically coherent masks. However, hierarchical variants also demonstrated increased recall over precision, implying increased false positives. The results demonstrate that explicit hierarchical structuring improves both performance and clinical plausibility, especially in low data dental imaging regimes. Code: \url{https://github.com/Banksylel/Restrictive-Hierarchical-Semantic-Segmentation}, Data: \url{https://zenodo.org/records/15038971}

\end{abstract}

\section{Introduction}

Within clinical practice, diseases are detected, staged and treatments planned in relation to anatomical structures within dental radiographs. Panoramic radiography provides a global view of the entire mouth, which reveals complex anatomical structures including teeth, surrounding alveolar bone, and other anatomical structures of the head and neck. Segmenting tooth and alveolar bone anatomies could improve the detection performance of dental caries and periodontal bone loss, while also providing a better understanding of dental scenes and precise localisation of said diseases. 

Dental anatomy naturally forms a hierarchical organisation, where high level classes such as ``tooth'' can be subdivided into finer classes like ``enamel'', ``dentin'', ``composite'', or ``pulp''. Within hierarchies such as this, coarser high level parent classes are generally easier to detect, whereas deeper classes are usually comprised of fine-grained featureless local features \cite{wu2019hierarchical}.

Recent works have explored the performance benefits of incorporating hierarchical class structures into segmentation frameworks. Muller et al. \cite{muller2020hierarchical} introduced a hierarchical loss that penalises misclassification according to individual detected child classes as well as their parent class. Li et al. \cite{li2022deep} developed HSSN, a framework that formulates semantic segmentation as multi-label classification over a structured class tree. Ke et al. \cite{ke2024learning} demonstrated hierarchical segmentation for recognition to boost task performance. Despite these works, most hierarchical segmentation models use a variation on hierarchy-aware loss that suggests child class weight adjustments based on its neighbouring child class performance as a summed parent class, or hierarchical feature fusion approach. This creates a limitation, where parent class masks can still have high false negative predictions with easy-to-detect features, due to child classes primarily being trained to detect low-level fine features. Additionally, using a loss function to incorporate hierarchical learning provides a non-definitive indirect detection of parent classes and does not leverage the easy-to-detect global features of parent classes \cite{banks2024hfcbformer}.

In this paper, we propose a novel deep learning approach for hierarchical semantic segmentation of dental structures in panoramic radiographs, that can be applied to any base model and data domain with coarse-to-fine feature data hierarchies. Our method introduces a hierarchy-aware, restrictive output framework that enables recurrent refinement of anatomical predictions across multiple hierarchy levels. By sequentially detecting coarse and fine classes within a structured hierarchy, and enforcing child predictions to be conditioned on parent detections. We demonstrate that our approach achieves improved segmentation performance and interpretability for stratified tooth layer and alveolar bone semantic segmentation.

\section{Methodology}

\subsection{Dataset}

\begin{figure}[!htbp]
    \centering
    \includegraphics[width=0.7\linewidth]{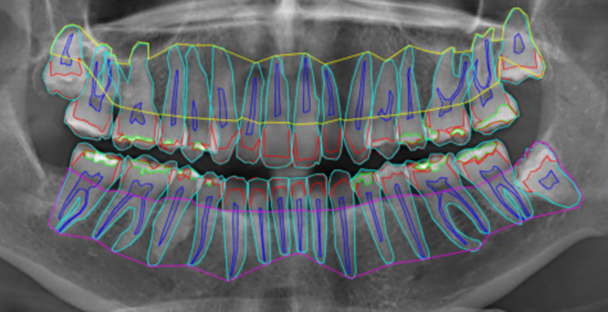}
    \caption{Cropped image of a panoramic radiograph with overlaid wireframe annotations for tooth layer dataset, of image 002.jpg. Classes include: Composite (green), Enamel (red), Pulp (dark blue), Tooth (light blue), Upper Alveolar Bone (yellow) and Lower Alveolar Bone (pink).}
    \label{fig:tl_annotatedimage}
\end{figure}

The proposed dataset, TL-pano \cite{banks2025tlpano}, consists of $194$ anonymous panoramic radiographs taken from $194$ patients with various backgrounds, ages, and dental hygiene practices, from Universidade de São Paulo, São Paulo, Brazil. Images were annotated with instance level multiclass and overlapping segmentation polygons, completed by three dental experts of at least $10$ years of experience each, using the VGG Image Annotator \cite{dutta2019vgg}. Images were annotated to indicate structural anatomies of the patients, such as tooth layers, FDI \cite{FDI2001} tooth numbers, composite material, and alveolar bone, an example of which is shown in Figure \ref{fig:tl_annotatedimage}. The majority of classes are annotated in a multiclass scheme, but the tooth layer class was given additional multilabel sub-classifications for FDI tooth numbering and quadrant numbering.

The collection of the original radiographs used for this dataset was approved by Universidade de São Paulo's ethics committee (ref: 5.681.738) and carried out with the appropriate patient informed consent procedure. 

\subsubsection{Data Processing}

For the purposes of this paper, the dataset was processed for a semantic segmentation setup without tooth numbering. The instance level objects are processed to follow a high to low priority order (composite, enamel, pulp, dentin, and alveolar bone). All object polygon instances are converted into pixel wise per object binary masks, checked for pixel overlap for higher priority classes, with the remaining distinct objects containing more than 50 pixels added to the final multiclass mask.

\begin{figure}[!htbp]
    \centering

    \begin{subfigure}[b]{0.48\linewidth}
        \centering
        \includegraphics[width=0.55\linewidth]{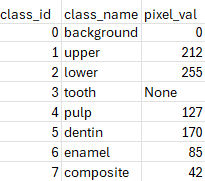}
        \caption{}
        \label{fig:classmap}
    \end{subfigure}
    \hfill
    \begin{subfigure}[b]{0.48\linewidth}
        \centering
        \includegraphics[width=0.65\linewidth]{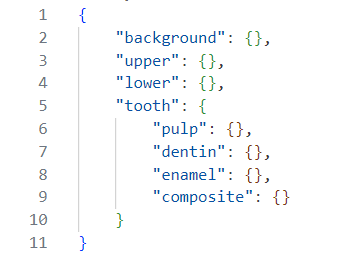}
        \caption{}
        \label{fig:classtree}
    \end{subfigure}

    \caption{Two images of the class\_map.CSV class to pixel map file (left) and the class\_tree.JSON hierarchy file (right).}
    \label{fig:classinfo}
\end{figure}

The target annotations are stored as a single uncompressed multiclass mask .PNG image for each radiograph, where each pixel value corresponds to a class when referencing the class\_map.CSV, shown in Figure \ref{fig:classmap}. The data is also setup for a hierarchy aware setting, where parent classes are considered the sum of their direct child classes. Class hierarchy information is stored in a class\_tree.JSON file, as seen in Figure \ref{fig:classtree}, where the class name is stored as the key, child classes are stored as the parent's values, with empty values for leaf nodes. class\_map.CSV and class\_tree.JSON are stored with the same character sensitive class name and in the same depth wise class order. Parent classes are not stored directly in the target masks, but instead should be generated by the data loader or evaluated as the sum of its child classes according to class\_tree.JSON.

\subsubsection{Classes and Counts}

For the semantic segmentation version of the dataset, we propose 7 positive classes, Upper Alveolar Bone, Lower Alveolar Bone, Tooth, Pulp, Dentin, Enamel and Composite. The Tooth class is the parent class of Pulp, Dentin, Enamel and Composite, so it is not directly stored in the mask.PNG files. Respective pixel and instance counts for each class are in Table \ref{tab:tl_clss_counts}.

\begin{table}[h!]
\centering
\scriptsize
\caption{Table containing class numbers per dataset, total instance counts and average pixel counts per image.}
\label{tab:tl_clss_counts}
\begin{tabular}{c|cccccccc|}
\cline{2-9}
& \multicolumn{8}{c|}{\textbf{Class Instance Counts}}                                                                                                                                       
\\ \cline{2-9} 
& \multicolumn{1}{c|}{\textbf{Background}} & \multicolumn{1}{c|}{\textbf{Composite}} & \multicolumn{1}{c|}{\textbf{Enamel}} & \multicolumn{1}{c|}{\textbf{Pulp}} & \multicolumn{1}{c|}{\textbf{Dentin}} & \multicolumn{1}{c|}{\textbf{Tooth}} & \multicolumn{1}{c|}{\textbf{Upper}} & \textbf{Lower} \\ \hline
\multicolumn{1}{|c|}{\textbf{Instance Num}} & \multicolumn{1}{c|}{-} & \multicolumn{1}{c|}{0}         & \multicolumn{1}{c|}{1}      & \multicolumn{1}{c|}{2}    & \multicolumn{1}{c|}{3}      & \multicolumn{1}{c|}{4}     & \multicolumn{1}{c|}{5}     & 6     \\ \hline
\multicolumn{1}{|c|}{\textbf{Instances}} &  \multicolumn{1}{c|}{-}   & \multicolumn{1}{c|}{1,328}      & \multicolumn{1}{c|}{5,873}   & \multicolumn{1}{c|}{5,392} & \multicolumn{1}{c|}{5,725}   & \multicolumn{1}{c|}{5,725}  & \multicolumn{1}{c|}{197}   & 197   \\ \hline
\multicolumn{1}{|c|}{\textbf{Semantic Num}} &  \multicolumn{1}{c|}{0}   & \multicolumn{1}{c|}{7}      & \multicolumn{1}{c|}{6}   & \multicolumn{1}{c|}{4} & \multicolumn{1}{c|}{5}   & \multicolumn{1}{c|}{3}  & \multicolumn{1}{c|}{1}   &  2  \\ \hline
\multicolumn{1}{|c|}{\textbf{Pixel P/Img}} &  \multicolumn{1}{c|}{5,161,360}   & \multicolumn{1}{c|}{38,744}      & \multicolumn{1}{c|}{147,300}   & \multicolumn{1}{c|}{99,225} & \multicolumn{1}{c|}{581,722}   & \multicolumn{1}{c|}{866,991}  & \multicolumn{1}{c|}{97,187}   &  159,646  \\ \hline
\end{tabular}
\end{table}

\begin{figure}[!htbp]
    \centering
    \begin{subfigure}[b]{0.24\textwidth}
        \centering
        \includegraphics[width=0.83\textwidth]{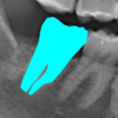}
        \caption{Tooth Class}
        \label{fig:tooth}
    \end{subfigure}
    \hfill
    \begin{subfigure}[b]{0.24\textwidth}
        \centering
        \includegraphics[width=0.78\textwidth]{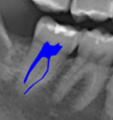}
        \caption{Pulp Class}
        \label{fig:pulp}
    \end{subfigure}
    \hfill
    \begin{subfigure}[b]{0.24\textwidth}
        \centering
        \includegraphics[width=0.8\textwidth]{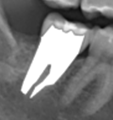}
        \caption{Dentin Class}
        \label{fig:dentin}
    \end{subfigure}
    
    \vspace{0.5cm}
    
    \begin{subfigure}[b]{0.24\textwidth}
        \centering
        \includegraphics[width=0.78\textwidth]{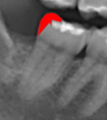}
        \caption{Enamel Class}
        \label{fig:enamel1}
    \end{subfigure}
    \hfill
    \begin{subfigure}[b]{0.24\textwidth}
        \centering
        \includegraphics[width=0.78\textwidth]{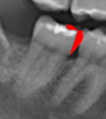}
        \caption{Enamel Class}
        \label{fig:enamel2}
    \end{subfigure}
    \hfill
    \begin{subfigure}[b]{0.24\textwidth}
        \centering
        \includegraphics[width=0.83\textwidth]{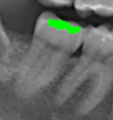}
        \caption{Composite}
        \label{fig:composite}
    \end{subfigure}
    
    \caption{Six diagrams containing the various classes for a single tooth, of image 002.}
    \label{fig:overall}
\end{figure}

The dataset proposes both tooth layer segmentation and alveolar bone segmentation tasks. Under the tooth layer segmentation task, the first class is the Tooth class, in Figure \ref{fig:tooth}, which indicates the entirety of each tooth as the sum of the Pulp, Dentin, Enamel and Composite classes. 

The Pulp class, in Figure \ref{fig:pulp} indicates the inner most layer of the tooth that contains the nerves and blood vessels. Pulp chambers with root canal treatments are considered Composite instead of Pulp. 

The Dentin class, in Figure \ref{fig:dentin}, is the middle layer between the Enamel and Pulp. The Dentin class for the instance version of the dataset is stored as an overlapping object with the pulp class, but the semantic segmentation version prioritises the Pulp class. 

The Enamel class, in Figure \ref{fig:enamel1} and Figure \ref{fig:enamel2}, indicate the outer most enamel layer of the tooth. Any class, such as the Enamel class, that has been split into two distinct objects due to decay or composite material, are treated as two distinct objects. 

The Composite class, in Figure \ref{fig:composite}, refers to any radiopaque artificial material within the teeth as a result of human intervention, and can be present in any layer of the tooth. Composite tooth material consists of fillings, caps, and root canals. Implants, bridges, and non-tooth related composites are excluded from this dataset.

\begin{figure}[!htbp]
    \centering
    \includegraphics[width=0.70\linewidth]{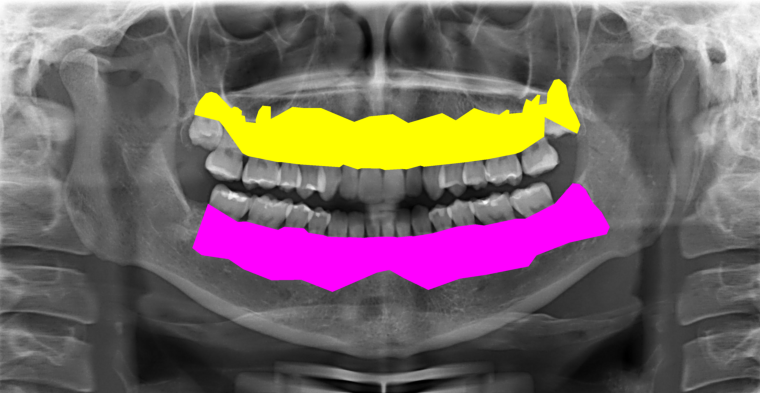}
    \caption{Diagram of a upper (yellow) and lower (pink) alveolar bone class annotations, of image 002.}
    \label{fig:alveolar}
\end{figure}

For non tooth annotations, the upper and lower alveolar bone class indicates the alveolar bone directly next to the teeth in the image, for the maxillary and mandibular jaws, as seen in Figure \ref{fig:alveolar}. The alveolar bone class was annotated to partially overlap with the teeth annotations, for ease of annotation. However, these were processed to remove the overlapping pixels from the alveolar bone classes, with priority given to the tooth classes.

\subsection{Hierarchy Setup}

\begin{figure}[!htbp]
    \centering
    \includegraphics[width=0.45\linewidth]{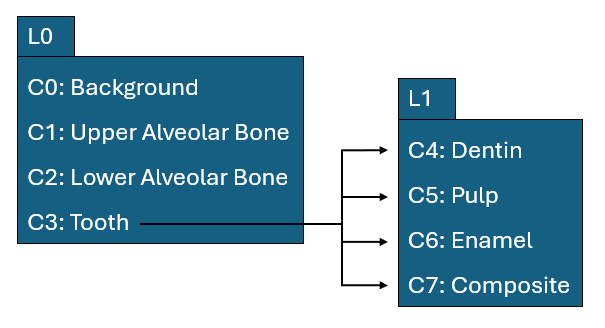}
    \caption{Image containing class hierarchy used with our dataset from the base nodes in Level 0 (L0) to the child classes in level 1 (L1)}
    \label{fig:hieararchy}
\end{figure}

The dataset classes are organised into a data hierarchy for a semantic segmentation setting, to facilitate the use of the proposed method, as seen in Figure \ref{fig:hieararchy}. Existing classes are ordered by injecting parent classes into the hierarchy, where the parent classes are made up only of its direct child classes. The class hierarchy is split into two levels, with one parent class in level 0 and four child classes in level 1. Parent classes are created under a set of proposed rules, (i) where there are more than two direct child classes available, (ii) there is a clear and distinct hierarchy where parent classes are made up of child classes, and (iii) a child class cannot exist over multiple branches. In a semantic segmentation setting, parent classes can be injected wherever the rules allow, even if classes of other branches bisect the objects of a child class into multiple objects. However, if implemented for an instance segmentation setting, branches must be distinct and contain all child classes within a single object. For the purposes of this work, hierarchy information is stored as a JSON file with nested fields indicating child classes, which correspond to a breadth wise ordered pixel mapping CSV file to determine leaf node pixel values.

\subsection{Hierarchical Model Design}

From a model design perspective, there are four primary elements, recurrent connections to feed features for a previous level into the next level with restrictive output nodes for each layer of the hierarchy, Feature-wise Linear Modulation (FiLM) feature conditioning \cite{perez2018film} to map high level features in deeper hierarchy levels, hierarchical probability composition to ensure all predictions follow the imposed hierarchical constraints, and hierarchical loss with weighted per level cross entropy and dice loss, with a hierarchy consistency loss to map marginal probabilities of parent classes to conditional child class probabilities. This method can be applied as a framework to any existing semantic segmentation model with minimal modifications. Our methodology is validated using 2 base donor models, with and without the restrictive hierarchical modifications. The two donor models adopted were chosen to evaluate the method with varying base architectures, UNet \cite{unet2015ronneberger} to evaluate with a relatively lower parameter fully convolutional network, and HRNet \cite{sun2019hrnet} to evaluate with a higher parameter network with fully connected layers.

\subsubsection{Recurrent Connections and Restrictive Output Nodes}

\begin{figure}[!htbp]
    \centering
    \includegraphics[width=1.0\linewidth]{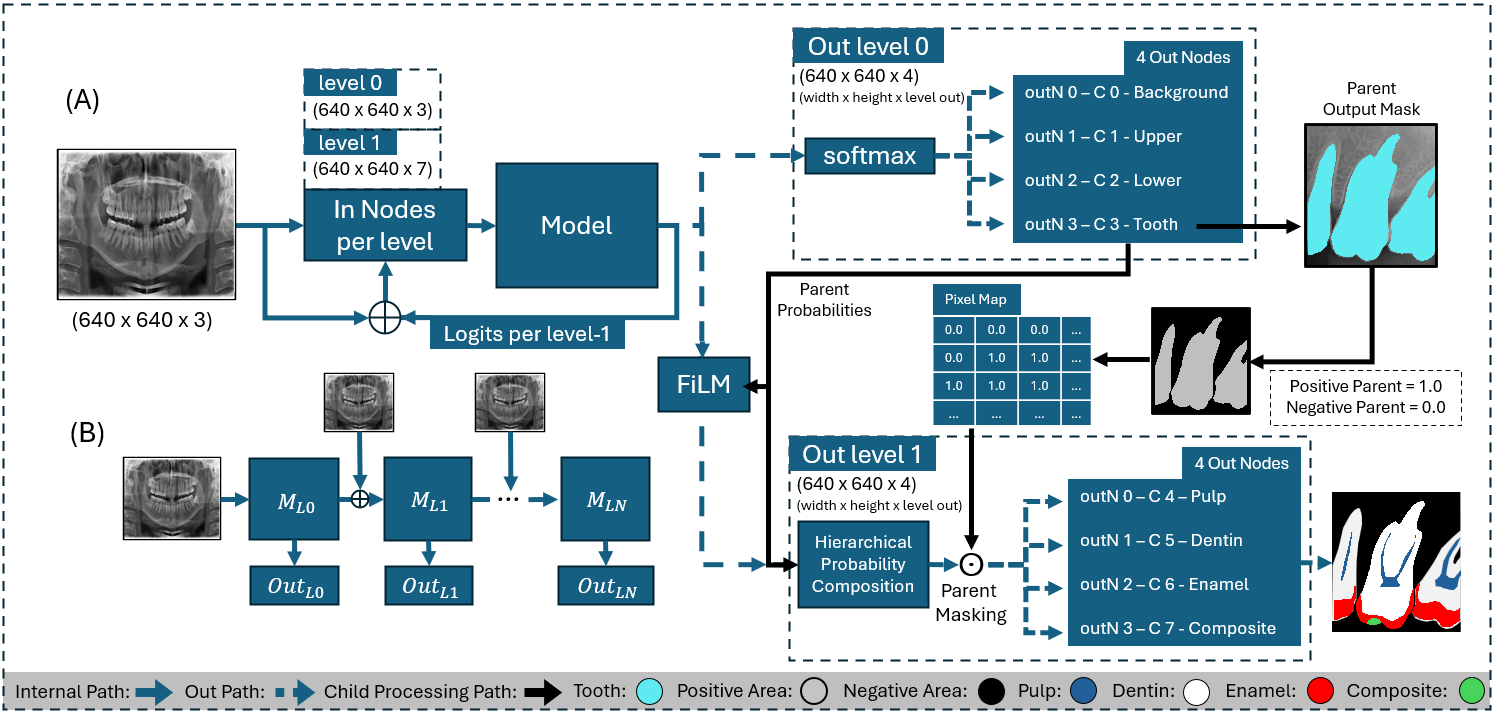}
    \caption{Diagram of the restrictive hierarchical modifications. (A) shows the model architecture including the recurrent connection and change in output and input per hierarchical level. (B) shows a linearised version spread out over $N$ hierarchy levels $L$, where model $M$ is the same model at each level with varying input/output nodes.}
    \label{fig:diagram}
\end{figure}

Our proposed method, in Figure \ref{fig:diagram}, is inspired by the recurrent connections of the recurrent neural networks \cite{rnn2020sherstinsky}. However, instead of feeding previous outputs with new information back into the model for each time step using the same output node shapes, it feeds the class logits of the previous hierarchy level along with the original input image back into the model and restricts the possible output nodes to only detect the member classes for a given hierarchy level.

The proposed hierarchical method takes the input image and feeds it into the donor model. This first pass through the model restricts the output nodes to only include the hierarchy level 0 classes in the output probability distribution. The pre-activation function logits are concatenated with the original input image and are fed back into the start of the model, now with a new set of input nodes that match the dimensions of the previous level's output nodes and input image. This new input is then fed through the model again, with the output nodes restricted to only include level 1 classes in the output probability distribution. This process is repeated until all classes are predicted by the model. All non-base hierarchy outputs (Level 1) are processed to remove predictions that are also not positively predicted in the previous layer's parent class for said child class, for both output logits and output level class probabilities. For non hierarchical baseline models, only leaf nodes are detected with a single pass though the model, with parent class evaluation being conducted on the combined child class masks.

Splitting the classes into a hierarchy and using recurrent connections per level of the hierarchy, the model can learn to retain coarse global features for the initial forward pass while refining outputs for finer features with each subsequent recurrent pass through the hierarchy. The recurrent passes back through the model follows an increasing coarse to fine feature refinement of the logits, closely following the coarse to fine hierarchy present in the dataset. Additionally, Concatenating the original input image at each hierarchy step ensures feature continuity, as only feeding the logits back would likely only retain global features after the initial first pass, disregarding the fine grained features early on.

Restricting the possible outputs of recurrent passes to only include the child classes for that hierarchy level (excluding background as an output for non-base level outputs), ensures the fine child class features within coarse parent class features can only be detected as one of the possible child class of that parent class, if already detected as its parent class in the previous forward pass. This creates a child class dependency on the performance of its direct parent class, but if the dataset follows a coarse to fine hierarchy, it could increase child class performance by leveraging the easy to detect coarse features of the parent classes.

\subsubsection{FiLM Feature Conditioning}

To enable deeper levels of the hierarchy to capture inter-level dependencies while preserving information extracted at coarser levels, the method applies FiLM feature conditioning to the per level output probabilities. Although FiLM was originally proposed to modulate static global embeddings within selected backbone layers using externally supplied conditioning information, our formulation repurposes FiLM to modulate dense, non-zero level child feature maps using the predicted probabilities of their direct parent classes.

In the hierarchical setting, the conditioning signal is not provided by an external modality. Instead, it is derived from the model's own predictions at the immediately preceding level of the hierarchy. For a single input image, the model produces a probability map at level $\ell-1$, consisting of one probability value for each class at each spatial location. To obtain a global conditioning vector, the probability values for each class are averaged across all spatial positions. This produces a vector whose dimensionality equals the number of classes at level $\ell-1$ and whose entries summarise the model's coarse level predictions for the entire image. This global vector is then passed through a shallow, level specific multilayer perceptron, which outputs the FiLM scaling and shifting parameters. Each of these parameter vectors has a length equal to the number of feature channels and provide a channel-wise modulation signal for the level-$\ell$ feature representation.

The resulting FiLM parameters are applied to the feature map corresponding to level $\ell$. Each channel of this feature map is scaled and shifted according to the FiLM outputs. This hierarchy conditioned FiLM module acts as a top down contextualisation mechanism, where the high level probability maps are condensed into a global summary vector, which then reshapes the backbone features used to classify fine level classes. Consequently, the backbone representation is dynamically reconfigured according to the model's current hierarchical and optimisation state.

\subsubsection{Hierarchical Probability Composition}
\label{sec:prob_composition}

Hierarchical probability composition is the mechanism through which the model enforces consistency between predictions made at different depths of the data hierarchy. It ensures that the probability assigned to any child class is dependent on the probability assigned to its parent. This hierarchy enforced chain rule, provides a method of propagating uncertainty from coarse to fine levels, preventing logically inconsistent predictions and enforcing the model to respect the underlying hierarchical relationships.

Each non-root class has a unique parent $p$, where $\mathcal{C}(p)$ is the parent's set of direct child classes. At the root level ($\ell = 0$), where classes do not share a structural dependency, the model applies an element wise sigmoid activation function to get base class probability distributions, which later act as a gating signal for subsequent hierarchy levels.

For each non-root level ($\ell > 0$), child-level logits are modified using the parent probability from the preceding level $p \in \ell-1$ using a spatially conditioned softmax \cite{koitka2024salt, memisevic2010gated},

\begin{equation}
Q^{(\ell)}(\mathcal{C}_{p} \mid p, x)
= \frac{\exp\left(Z^{(\ell)}(\mathcal{C}_{p} \mid x) + \log\left(P^{(\ell-1)}(p \mid x) + \epsilon\right)\right)}
{\displaystyle\sum_{c \in \mathcal{C}_{p}} \exp\left(Z^{(\ell)}(c \mid x) 
+ \log\left(P^{(\ell-1)}(p \mid x) + \epsilon\right)\right)},
\label{eq:softmax_hier2}
\end{equation}

\noindent where $x$ is the input image, $Z^{(\ell)} \in \mathbb{R}^{|\mathcal{C}_\ell(p)| \times W \times H}$ are the logits corresponding to the parent's child class group, and $\epsilon$ is a small constant. This determines the probability distribution of the children classes based on the probability of its parent.

The final absolute probability for each child is obtained by composing the parent's probability map with the child conditional distribution, 
\begin{equation}
P^{(\ell)}(\mathcal{C}_{p} \mid x)
= P^{(\ell-1)}(p \mid x)\, Q^{(\ell)}(\mathcal{C}_{p} \mid p, x).
\label{eq:prob_composition}
\end{equation}
\noindent This enforces the hierarchical constraint $P^{(\ell)}(c \mid x) \leq P^{(\ell-1)}(p \mid x),$ ensuring a child class can never receive a higher probability than its parent, by suppressing all child logits if the parent is unlikely and amplifying them when the parent is confident. This also allows for uncertainty to propagate downward, so low confidence in parent classes induce low confidence in all of its descendants. 

A potential concern with conditional probability models is within deep hierarchies, uncertainty at higher levels may accumulate and suppress the probabilities of deep child classes. The proposed method mitigates this by ensuring each per parent group of children classes are normalised with softmax, ensuring that the conditional distribution always sums to one. This prevents degeneracy within each child class group and encourages the optimiser to form a confident path down the hierarchy for the correct structure. Additionally, FiLM conditioning injects parent-level probabilities into the feature representation for the next level, allowing strong child class probabilities to back-propagate and correct uncertain parent prediction. This stabilises the chain rule composition and prevents deep child probabilities from vanishing even in deep hierarchies.

\subsubsection{Hierarchical Loss Functions}

To enable loss across hierarchical levels within the model, two primary loss functions are utilised, a per level weighted combination dice and cross entropy loss to independently capture level specific loss, and a hierarchical consistency loss to map coherence between parent and child class probabilities.

\noindent Weighted hierarchical loss functions are necessary to capture per level losses along with the learning benefit given by the original loss functions. However, direct use of the loss functions on each level won't appropriately capture per level loss. Instead, child class losses ignore pixels that are also not in their direct parent class from calculation, creating a child class loss that only considers the area of its direct parent class, naturally weighting loss in favour of child classes. An inverse median frequency weighted loss \cite{eigen2015predicting} scheme is also incorporated, to handle the heavy class imbalance present within the dataset. The loss functions contain hierarchical version of weighted cross entropy and dice loss, 

\begin{equation}
\mathcal{L}_{\mathrm{Dice}}^{(\ell)} = 1 - \frac{1}{C_\ell} \sum_{c}^{C_\ell}
\frac{2 \sum_{x} m_c(x) w_c y^{(\ell)}_c(x) P^{(\ell)}_c(x) + \epsilon}
{\sum_{x} m_c(x) w_c( y^{(\ell)}_c(x) + P^{(\ell)}_c(x) ) + \epsilon},
\label{eq:hier_loss_dice}
\end{equation}

\begin{equation}
\mathcal{L}_{\mathrm{CE}}^{(\ell)} = - \frac{1}{C_\ell} \sum_{c}^{C_\ell} \sum_{x} m_c(x) w_c 
y^{(\ell)}_c(x) \log \left( P^{(\ell)}_c(x) \right).
\label{eq:hier_loss_ce}
\end{equation}

\begin{equation}
\mathcal{L}_{\mathrm{WHL}} = \sum_{\ell} \mathcal{L}_{\mathrm{CE}}^{(\ell)} + \mathcal{L}_{\mathrm{Dice}}^{(\ell)}.
\label{eq:hier_loss_combined}
\end{equation}

\noindent where $P^{(\ell)}_c(x)$ is the output probability for a given class, $y^{(\ell)}_c(x) \in \{0,1,-1\}$ is the one hot encoded target, $w_c$ is the class weight value, and $m_c(x) \in \{0,1\}$ is a parent visibility map. The parent visibility map indicates which pixels are positively indicated by the parent class as $1.0$, and which are not indicated by the parent $0.0$. Base level classes have a parent visibility map of $1.0$ for all pixels.

\noindent While the hierarchical composition mechanism already encodes the dependencies between classes, unconstrained optimisation can still produce temporary inconsistencies between parent and child activations. This is particularly apparent in early optimisation steps, where poor parent predictions can suppress the optimisation signal for their descendants. To mitigate this, Hierarchical consistency loss is introduced to penalise deviations from the condition that each parent's marginal probability should equal the sum of its children probabilities.

The comparison between parent and direct child class probabilities is defined as the difference between the sum of child class probabilities and the parent class probability averaged over all valid parent child groups and levels,
\begin{equation}
\mathcal{L}_{\mathrm{HC}} = \frac{\sum_{\ell} \sum_{p}\left| \sum_{c \in \mathcal{C}_{p}} P^{(\ell)}(c \mid x) - P^{(\ell-1)}(p \mid x) \right|}{NP}.
\label{eq:consistency_loss1}
\end{equation}

\noindent where $NP$ is the total number of parent sets in the hierarchy. 

The final loss function combines the supervised weighted hierarchical and hierarchical consistency loss,
\begin{equation}
\mathcal{L} = \mathcal{L}_{\mathrm{WHL}} + \mathcal{L}_{\mathrm{HC}}.
\label{eq:consistency_loss3}
\end{equation}

\section{Results}

\subsection{Experiment Setup and Donor Model Choice}

To evaluate the proposed method, this section compares the hierarchical and non hierarchical model semantic segmentation performance for two donor models, UNet and HRNet. The total parameters for each model are $13,395,719$ (UNet), $65,850,727$ (HRNet), $13,396,424$ (UNet-H), $65,858,648$ (UNet-H). 

From the original 197 image mask pairs, 19 (10$\%$) are randomly sampled as a hold out test set. Out of the remaining 178 (90$\%$) image mask pairs, the pairs are split into 5 folds of 35/36 validation pairs (20$\%$), where for each validation split the remaining unused 143/142 pairs (80$\%$) are used for training. All folds are pre split and all models are trained and evaluated the same splits. All images are greyscale and trained as 1 channel images. In this low data regime, each of the models are initialised using their respective pretrained weights on an out of distribution dataset, before fine tuning on the proposed dataset.

For experimental consistency, all models utilise a base weighted loss function of dice loss and cross entropy loss, with additional hierarchical losses for the hierarchical versions. Weighted factors for the loss functions are determined using an inverse median frequency scheme for each of the detected outputs, hierarchical weights are in Equation \ref{eq:weight_loss1} and non hierarchical weights are in Equation \ref{eq:weight_loss2}. Each method uses an independently tuned starting learning rate, optimised with a one-dimensional grid search, where the learning rate reduces by a factor of 0.5 ($lr\times0.5$) if the loss does not improve for 3 epochs with a minimum of 0.001. Starting learning rates for each model are: UNet, 0.018; HRNet, 0.022; UNet-H, 0.022; HRNet-H, 0.024. Each model has a total number of epochs of 80, where the best weights are saved for evaluation, determined by validation IoU performance. Each model has a training batch size of 4, image size of $640\times640$ pixels, and the AdamW optimiser. An NVIDIA RTX 3090 GPU and 64GB of RAM is used for training. 

\begin{equation}
\begin{aligned}
    \ell0[c0=0.0297, c1=1.577, c2=0.9619, c3=0.1770],\\ 
    \ell1[c4=1.5432, c5=0.2638, c6=1.0413, c7=3.9722]
    \label{eq:weight_loss1}
\end{aligned}
\end{equation}
\begin{equation}
    [c0=0.0285, c1=1.5159, c2=0.9227, c4=1.4842, c5=0.2532, c6=1.0, c7=3.8021]
    \label{eq:weight_loss2}
\end{equation}

All models use the same light augmentation, due to low variation between object positions in panoramic radiographs. The models use: Gaussian blur with a kernel of (25, 25) and sigma of (0.001, 0.2); jitter of brightness 0.4, contrast of 0.5, saturation of 0.25, and hue of 0.01; horizontal flip of probability 0.5; no vertical flip; affine translation of probability 1.0, rotation of (-50, 50), horizontal translation of (-20, 20), vertical translation of (-20,20), scale of (0.85, 1.15), and shear of (-5, 5).

Model performance is evaluated using semantic segmentation metrics at predicting each class. Metrics include IoU = $\frac{TP}{TP+FP+FN}$, Dice = $\frac{2TP}{2TP+FP+FN}$, Precision = $\frac{TP}{TP+FP}$, Recall =  $\frac{TP}{TP+FN}$. True Positive (TP), True Negative (TN), False Positive (FP), and False Negative (FN) are the numbers of pixels in an image for a specific class. Final metrics are presented as the mean (standard deviation) metrics for all validation and test images in the dataset using a 5 fold cross validation scheme. The standard deviation is calculated over each of the 5 folds. For hierarchical methods that mask some child class target pixels as -1 for loss calculation, if the pixel is also not positive for the direct parent class, the -1 pixels are converted to 0 for fair evaluation of the full image. Hierarchical model metrics are performed on the direct model outputs, while non hierarchical model parent class outputs are processed to use the sum of their child class outputs. 

Results are presented primarily as validation set performance, with test set performance placed in the Appendix \ref{app:1}.

\subsection{Quantitative Results}

Comparing the performance of the donor models with and without the proposed hierarchical methodology, in Table \ref{tab:hier_seg2_results_val}, the hierarchical methodology consistently improved overall IoU, Dice and Recall segmentation performance for child classes, on the validation set for both models. However, for UNet-H, this gain in performance for child classes results in a slight reduction in performance for base, parent, and Pulp classes, likely due to narrow bottlenecks present in UNet and loss function that favours child class performance. HRNet-H in comparison, shows a consistent increase in performance for all classes with IoU, Dice and Recall. Both models also exhibit a reduction in Precision for certain classes, implying parent class areas are over predicted to increase the possible detected area for child class dependencies, slightly reducing Precision for all classes as a result.

\begin{table}[hbt!]
\centering
\renewcommand{\arraystretch}{1.}
\caption{Per-class semantic segmentation metrics mean ($\pm$ standard deviation), comparing Original and Hierarchical versions of UNet and HRNet on the validation set. Bold values are better performing when comparing original and hierarchical versions.}
\resizebox{\textwidth}{!}{
\begin{tabular}{|ll|cccc|cccc|}
\toprule
\multicolumn{2}{|c|}{} & \multicolumn{8}{c|}{\textbf{Validation Set}} \\
\cmidrule(lr){3-10}
\multicolumn{2}{|c}{} & \multicolumn{4}{|c}{\textbf{Without Hierarchy}} & \multicolumn{4}{|c|}{\textbf{With Hierarchy}} \\
\cmidrule(lr){3-6} \cmidrule(lr){7-10}
\textbf{Model} & \textbf{Class} & IoU & Dice & Precision & Recall & IoU & Dice & Precision & Recall \\
\midrule
\multirow{9}{*}{\textbf{UNet}} & Average & 0.692 ($\pm$ 0.009) & 0.792 ($\pm$ 0.009) & \textbf{0.802} ($\pm$ 0.008) & 0.795 ($\pm$ 0.012) & \textbf{0.696} ($\pm$ 0.007) & \textbf{0.794} ($\pm$ 0.007) & 0.782 ($\pm$ 0.008) & \textbf{0.819} ($\pm$ 0.010) \\
& Background & \textbf{0.980} ($\pm$ 0.001) & \textbf{0.990} ($\pm$ 0.000) & \textbf{0.991} ($\pm$ 0.001) & \textbf{0.989} ($\pm$ 0.001) & 0.979 ($\pm$ 0.001) & 0.989 ($\pm$ 0.001) & 0.990 ($\pm$ 0.001) & 0.988 ($\pm$ 0.001) \\
& Upper & \textbf{0.535} ($\pm$ 0.011) & \textbf{0.694} ($\pm$ 0.010) & \textbf{0.669} ($\pm$ 0.033) & 0.732 ($\pm$ 0.029) & 0.523 ($\pm$ 0.007) & 0.684 ($\pm$ 0.006) & 0.640 ($\pm$ 0.012) & \textbf{0.745} ($\pm$ 0.010) \\
& Lower & \textbf{0.716} ($\pm$ 0.006) & \textbf{0.833} ($\pm$ 0.004) & \textbf{0.820} ($\pm$ 0.006) & 0.850 ($\pm$ 0.008) & 0.697 ($\pm$ 0.012) & 0.820 ($\pm$ 0.009) & 0.791 ($\pm$ 0.017) & \textbf{0.856} ($\pm$ 0.016) \\
& Tooth & \textbf{0.860} ($\pm$ 0.006) & \textbf{0.924} ($\pm$ 0.003) & 0.926 ($\pm$ 0.005) & \textbf{0.924} ($\pm$ 0.011) & 0.850 ($\pm$ 0.005) & 0.919 ($\pm$ 0.003) & \textbf{0.928} ($\pm$ 0.007) & 0.911 ($\pm$ 0.007) \\
& Pulp & \textbf{0.579} ($\pm$ 0.010) & \textbf{0.731} ($\pm$ 0.008) & \textbf{0.731} ($\pm$ 0.010) & 0.742 ($\pm$ 0.011) & 0.566 ($\pm$ 0.009) & 0.719 ($\pm$ 0.007) & 0.673 ($\pm$ 0.013) & \textbf{0.787} ($\pm$ 0.010) \\
& Dentin & 0.776 ($\pm$ 0.013) & 0.873 ($\pm$ 0.009) & 0.919 ($\pm$ 0.006) & 0.833 ($\pm$ 0.012) & \textbf{0.853} ($\pm$ 0.006) & \textbf{0.920} ($\pm$ 0.003) & \textbf{0.936} ($\pm$ 0.003) & \textbf{0.905} ($\pm$ 0.008) \\
& Enamel & 0.724 ($\pm$ 0.037) & 0.828 ($\pm$ 0.030) & \textbf{0.831} ($\pm$ 0.028) & 0.839 ($\pm$ 0.021) & \textbf{0.735} ($\pm$ 0.035) & \textbf{0.834} ($\pm$ 0.030) & 0.819 ($\pm$ 0.025) & \textbf{0.864} ($\pm$ 0.031) \\
& Composite & \textbf{0.368} ($\pm$ 0.066) & 0.463 ($\pm$ 0.080) & \textbf{0.529} ($\pm$ 0.070) & 0.452 ($\pm$ 0.100) & 0.368 ($\pm$ 0.056) & \textbf{0.463} ($\pm$ 0.065) & 0.482 ($\pm$ 0.043) & \textbf{0.498} ($\pm$ 0.108) \\
\midrule
\multirow{9}{*}{\textbf{HRNet}} & Average & 0.656 ($\pm$ 0.010) & 0.765 ($\pm$ 0.011) & \textbf{0.772} ($\pm$ 0.007) & 0.772 ($\pm$ 0.016) & \textbf{0.679} ($\pm$ 0.005) & \textbf{0.781} ($\pm$ 0.006) & 0.769 ($\pm$ 0.006) & \textbf{0.808} ($\pm$ 0.007) \\
& Background & 0.977 ($\pm$ 0.001) & 0.988 ($\pm$ 0.000) & 0.989 ($\pm$ 0.001) & 0.988 ($\pm$ 0.001) & \textbf{0.979} ($\pm$ 0.001) & \textbf{0.989} ($\pm$ 0.000) & \textbf{0.990} ($\pm$ 0.001) & \textbf{0.988} ($\pm$ 0.001) \\
& Upper & 0.521 ($\pm$ 0.014) & 0.682 ($\pm$ 0.012) & \textbf{0.651} ($\pm$ 0.014) & 0.727 ($\pm$ 0.014) & \textbf{0.529} ($\pm$ 0.011) & \textbf{0.689} ($\pm$ 0.010) & 0.648 ($\pm$ 0.014) & \textbf{0.745} ($\pm$ 0.014) \\
& Lower & 0.690 ($\pm$ 0.009) & 0.815 ($\pm$ 0.007) & \textbf{0.805} ($\pm$ 0.008) & 0.830 ($\pm$ 0.015) & \textbf{0.701} ($\pm$ 0.003) & \textbf{0.823} ($\pm$ 0.002) & 0.800 ($\pm$ 0.010) & \textbf{0.851} ($\pm$ 0.008) \\
& Tooth & 0.838 ($\pm$ 0.004) & 0.912 ($\pm$ 0.003) & 0.915 ($\pm$ 0.005) & 0.909 ($\pm$ 0.006) & \textbf{0.847} ($\pm$ 0.004) & \textbf{0.917} ($\pm$ 0.002) & \textbf{0.924} ($\pm$ 0.003) & \textbf{0.910} ($\pm$ 0.005) \\
& Pulp & 0.480 ($\pm$ 0.014) & 0.646 ($\pm$ 0.012) & \textbf{0.624} ($\pm$ 0.011) & 0.685 ($\pm$ 0.029) & \textbf{0.496} ($\pm$ 0.012) & \textbf{0.660} ($\pm$ 0.011) & 0.593 ($\pm$ 0.022) & \textbf{0.763} ($\pm$ 0.009) \\
& Dentin & 0.722 ($\pm$ 0.009) & 0.838 ($\pm$ 0.006) & 0.903 ($\pm$ 0.010) & 0.783 ($\pm$ 0.008) & \textbf{0.817} ($\pm$ 0.006) & \textbf{0.899} ($\pm$ 0.003) & \textbf{0.930} ($\pm$ 0.002) & \textbf{0.871} ($\pm$ 0.008) \\
& Enamel & 0.670 ($\pm$ 0.030) & 0.791 ($\pm$ 0.026) & \textbf{0.797} ($\pm$ 0.027) & 0.798 ($\pm$ 0.030) & \textbf{0.709} ($\pm$ 0.030) & \textbf{0.817} ($\pm$ 0.026) & 0.793 ($\pm$ 0.028) & \textbf{0.857} ($\pm$ 0.021) \\
& Composite & 0.347 ($\pm$ 0.058) & 0.445 ($\pm$ 0.070) & \textbf{0.492} ($\pm$ 0.062) & 0.454 ($\pm$ 0.089) & \textbf{0.354} ($\pm$ 0.052) & \textbf{0.453} ($\pm$ 0.062) & 0.474 ($\pm$ 0.044) & \textbf{0.482} ($\pm$ 0.090) \\
\bottomrule
\end{tabular}
}
\label{tab:hier_seg2_results_val}
\end{table}

Analysing specific results for UNet, the hierarchical model produced a modest yet reliable increase in average IoU of 0.696 compared to 0.692 and Dice of 0.793 compared to 0.792, as a result of improved recall across nearly all non background classes. Comparably, child classes benefit substantially with an average child IoU increase from 0.611 to 0.631 and Dice from 0.724 to 0.734. HRNet However, exhibited consistent increases in performance for all classes, with an average IoU increase from 0.656 to 0.679 and Dice from 0.765 to 0.781, with the largest relative improvements again concentrated in child classes with an average IoU increase of $+0.039$, compared to parent and base class increase of $+0.007$ IoU. These improvements were achieved without degradation in background segmentation performance, indicating that the hierarchical mechanism does not introduce overfitting to foreground structures or compromise global consistency.

\subsection{Qualitative Results}

\begin{figure}[!htbp]
    \centering
    \begin{subfigure}[b]{0.33\textwidth}
        \centering
        \includegraphics[width=\textwidth]{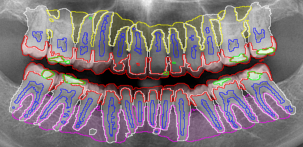}
        \caption{UNet Image 043}
        \label{fig:results_hier2_1}
    \end{subfigure}
    \hfill
    \begin{subfigure}[b]{0.28\textwidth}
        \centering
        \includegraphics[width=\textwidth]{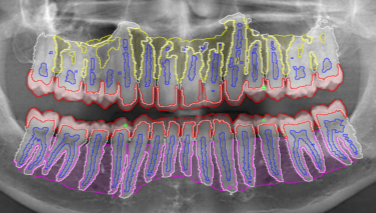}
        \caption{UNet Image 112}
        \label{fig:results_hier2_2}
    \end{subfigure}
    \hfill
    \begin{subfigure}[b]{0.36\textwidth}
        \centering
        \includegraphics[width=\textwidth]{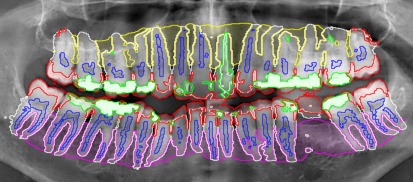}
        \caption{UNet Image 189}
        \label{fig:results_hier2_3}
    \end{subfigure}

    \vspace{0.2cm}

    \begin{subfigure}[b]{0.31\textwidth}
        \centering
        \includegraphics[width=\textwidth]{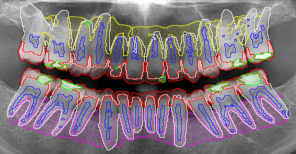}
        \caption{UNet-H Image 043}
        \label{fig:results_hier2_4}
    \end{subfigure}
    \hfill
    \begin{subfigure}[b]{0.3\textwidth}
        \centering
        \includegraphics[width=\textwidth]{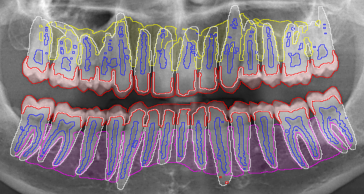}
        \caption{UNet-H Image 112}
        \label{fig:results_hier2_5}
    \end{subfigure}
    \hfill
    \begin{subfigure}[b]{0.36\textwidth}
        \centering
        \includegraphics[width=\textwidth]{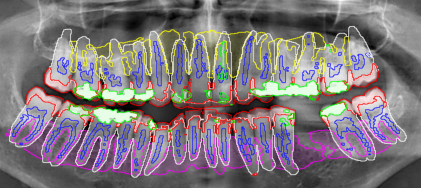}
        \caption{UNet-H Image 189}
        \label{fig:results_hier2_6}
    \end{subfigure}

    \vspace{0.2cm}

    \begin{subfigure}[b]{0.31\textwidth}
        \centering
        \includegraphics[width=\textwidth]{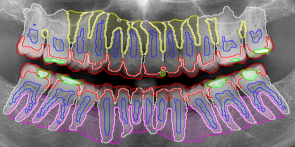}
        \caption{HRNet Image 043}
        \label{fig:results_hier2_7}
    \end{subfigure}
    \hfill
    \begin{subfigure}[b]{0.31\textwidth}
        \centering
        \includegraphics[width=\textwidth]{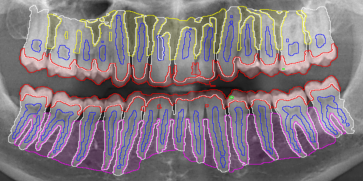}
        \caption{HRNet Image 112}
        \label{fig:results_hier2_8}
    \end{subfigure}
    \hfill
    \begin{subfigure}[b]{0.34\textwidth}
        \centering
        \includegraphics[width=\textwidth]{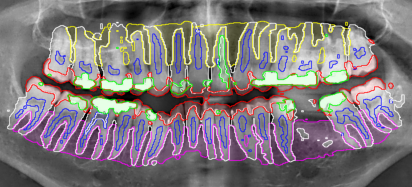}
        \caption{HRNet Image 189}
        \label{fig:results_hier2_9}
    \end{subfigure}

    \vspace{0.2cm}

    \begin{subfigure}[b]{0.31\textwidth}
        \centering
        \includegraphics[width=\textwidth]{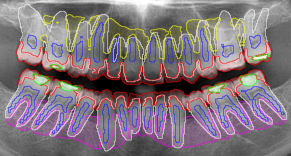}
        \caption{HRNet-H Image 043}
        \label{fig:results_hier2_10}
    \end{subfigure}
    \hfill
    \begin{subfigure}[b]{0.31\textwidth}
        \centering
        \includegraphics[width=\textwidth]{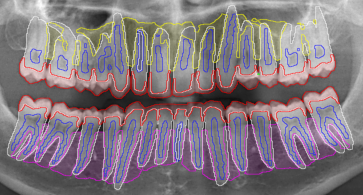}
        \caption{HRNet-H Image 112}
        \label{fig:results_hier2_11}
    \end{subfigure}
    \hfill
    \begin{subfigure}[b]{0.35\textwidth}
        \centering
        \includegraphics[width=\textwidth]{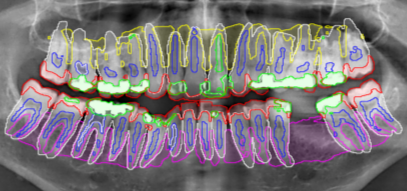}
        \caption{HRNet-H Image 189}
        \label{fig:results_hier2_12}
    \end{subfigure}

    \vspace{0.2cm}

    \begin{subfigure}[b]{0.31\textwidth}
        \centering
        \includegraphics[width=\textwidth]{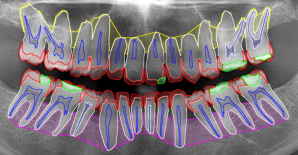}
        \caption{Target Image 043}
        \label{fig:results_hier2_13}
    \end{subfigure}
    \hfill
    \begin{subfigure}[b]{0.3\textwidth}
        \centering
        \includegraphics[width=\textwidth]{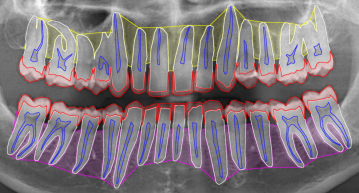}
        \caption{Target Image 112}
        \label{fig:results_hier2_14}
    \end{subfigure}
    \hfill
    \begin{subfigure}[b]{0.37\textwidth}
        \centering
        \includegraphics[width=\textwidth]{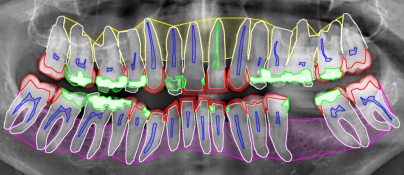}
        \caption{Target Image 189}
        \label{fig:results_hier2_15}
    \end{subfigure}
    
    \caption{Figure containing images with overlaid predictions and targets, for each model. Classes include: Upper Alveolar Bone (yellow), Lower Alveolar Bone (pink), Pulp (dark blue), Dentin (white), Enamel (red) and Composite (green).}
    \label{fig:results_hier2}
\end{figure}

Evaluating the method qualitatively, as seen in Figure \ref{fig:results_hier2}, all models exhibit overall well performing predictions at most segmentation tasks, but with occasional poor performance for certain edge case samples. Additionally, while all models perform well at detecting the majority of the classes in each image, the quality of the predicted edges are often noisy. Generally, HRNet-H performed best followed by UNet-H, HRNet and UNet. Non hierarchical versions of HRNet and UNet often exhibit noisier predictions when compared to their hierarchical counterparts.

A common false positive prediction present in non hierarchical versions is a large number of independent floating predictions, such as in Figure \ref{fig:results_hier2_1}, \ref{fig:results_hier2_3} and \ref{fig:results_hier2_9}, there are spots of dentin predictions away from the tooth within the alveolar bone. These false predictions are not present in the hierarchical versions, as each fine grained child prediction is constrained by its parent class, ensuring that detailed features in the dentin class is only detected where the broader easily detected tooth class is also present. However, occasionally this behaviour can still exist within the bounds of the parent class, such as within Figure \ref{fig:results_hier2_5} which contain enamel predictions at the end of a root.

All models produce predictions with a large number of small split object for classes with unclear boundaries, such as pulp and tooth predictions in Figure \ref{fig:results_hier2_2} and \ref{fig:results_hier2_12}. Although, this prediction behaviour is more prevalent in the non hierarchical versions, with HRNet-H producing the least of these errors.

Image 112 depicts an example of the average performance for each model. However, as seen in image 189, certain edge cases such as missing teeth, impacted third molars, or extensive reconstructive work, can cause the models to perform poorly for those samples. However, even though all models perform poorly on these edge case samples, both hierarchical versions generally perform better than their non hierarchical counterparts. All models generally perform well at alveolar bone segmentation, but their performance is directly determined by the predicted quality of the other nearby classes. Dentin predictions that have an over predicted area, like in Figure \ref{fig:results_hier2_1} and \ref{fig:results_hier2_9}, reduce the performance of the alveolar bone classes. Although, some models also struggle to detect full alveolar bone areas that contain missing teeth, as seen in Figure \ref{fig:results_hier2_6}.

Hierarchical versions of both models exhibit higher quality predictions for almost all cases. While quantitative gain is relatively mediocre, qualitative gain is heavily improved as they contain small improvements with much more cohesive predictions, increasing the clinical significance of the hierarchical models.

\section{Conclusion}

We propose a method of mapping and learning inherent data hierarchies represented in a proposed dental anatomy radiographic dataset, for use with deep learning based multiclass semantic segmentation models. The method utilises a recurrent neural network inspired structure with a variable class output at each hierarchy step and a child class output formulation, where the model is forced to detect one of the possible child class outputs for the area also positively detected by their parent. We utilise a FiLM feature conditioning and proposed hierarchical probability composition that retains differentiable parent to child class feature probabilities. Model optimisation is achieved by calculating individual class performance as well as parent to child class hierarchical performance. 

Our method exhibits performance increases for all classes for higher parameter models utilising fully connected layers, but is increased only for child classes at the expense of parent classes for lower parameter fully convolutional models. However, utilising the proposed method for both donor model does also reduce Precision and boost Recall, implying increased false positives and a tendency to over predict per class locations. From a clinical point of view, despite the very small 142 to 143 image training set sizes, performance is high both quantitatively and qualitatively, producing anatomically plausible predictions. Both hierarchical and non hierarchical models produce clinically useful predictions with some models performing better than others, of which the best performing model is UNet-H. This high baseline performance is likely due to the usually well defined boundaries between individual tooth layers and alveolar bone, allowing for an easier detection task. However, there are cases where alveolar bone and tooth boundaries are unclear, which is where the hierarchical formulations exhibit clear improvement. Additionally, the hierarchical formulations utilise easy to detect parent dependencies for child class predictions, reducing false predictions in clearly incorrect locations, such as dentin being detected in alveolar bone locations with missing teeth.

The dataset size raises concerns about true performance and real world performance, although similar cross validation and test set performance with low standard deviation still implies model generalisability. Despite clinically useful predictions, the detection task of tooth layer segmentation is not overly significant on its own. However, if used in conjunction with additional disease datasets, it could improve a model's understanding of the domain, taking a step towards more intelligent detection methodologies that could improve real world clinical processes, such as automated charting. Specifically, utilising this dataset and hierarchical methodology in conjunction with additional caries segmentation imbedded into the existing hierarchy, could vastly improve detection of the comparatively much lower features and object sizes represented in caries lesions. Additionally, the detection of tooth layers in addition to caries lesions could provide a definitive staging and classification method for caries lesions, determined by their presence within certain layers rather than subtle feature differences between disease stages. Future work should explore methods of incorporating distributed external caries detection datasets within existing tooth layer hierarchies, increasing annotated image counts, and exploring high depth hierarchies with potentially improved methodologies to map hierarchical dependencies throughout entire long hierarchy chains.

\bibliographystyle{ieeetr}
\bibliography{refs.bib}

\section*{Appendix}
\label{app:1}

Test set performance is shown in Table \ref{tab:test_results}. They show similar, but slightly increased, overall performance when compared to the validation set performance. 

\begin{table}[hbt!]
\centering
\renewcommand{\arraystretch}{1.}
\caption{Per-class semantic segmentation metrics mean ($\pm$ standard deviation), comparing Original and Hierarchical versions of UNet and HRNet on the test set. Bold values are better performing when comparing original and hierarchical versions.}
\resizebox{\textwidth}{!}{
\begin{tabular}{|ll|cccc|cccc|}
\toprule
\multicolumn{2}{|c|}{} & \multicolumn{8}{c|}{\textbf{Test Set}} \\
\cmidrule(lr){3-10}
\multicolumn{2}{|c}{} & \multicolumn{4}{|c}{\textbf{Without Hierarchy}} & \multicolumn{4}{|c|}{\textbf{With Hierarchy}} \\
\cmidrule(lr){3-6} \cmidrule(lr){7-10}
\textbf{Model} & \textbf{Class} & IoU & Dice & Precision & Recall & IoU & Dice & Precision & Recall \\
\midrule
\multirow{9}{*}{\textbf{UNet}} & Average & 0.706 ($\pm$ 0.007) & 0.805 ($\pm$ 0.006) & \textbf{0.815} ($\pm$ 0.006) & 0.805 ($\pm$ 0.004) & \textbf{0.709} ($\pm$ 0.008) & \textbf{0.806} ($\pm$ 0.006) & 0.793 ($\pm$ 0.009) & \textbf{0.828} ($\pm$ 0.004) \\
& Background & \textbf{0.981} ($\pm$ 0.001) & \textbf{0.990} ($\pm$ 0.000) & \textbf{0.992} ($\pm$ 0.001) & \textbf{0.988} ($\pm$ 0.001) & 0.979 ($\pm$ 0.001) & 0.990 ($\pm$ 0.001) & 0.992 ($\pm$ 0.001) & 0.988 ($\pm$ 0.001) \\
& Upper & \textbf{0.528} ($\pm$ 0.008) & \textbf{0.689} ($\pm$ 0.007) & \textbf{0.638} ($\pm$ 0.024) & 0.758 ($\pm$ 0.029) & 0.515 ($\pm$ 0.007) & 0.678 ($\pm$ 0.006) & 0.615 ($\pm$ 0.004) & \textbf{0.764} ($\pm$ 0.014) \\
& Lower & \textbf{0.737} ($\pm$ 0.007) & \textbf{0.848} ($\pm$ 0.005) & \textbf{0.839} ($\pm$ 0.012) & 0.859 ($\pm$ 0.005) & 0.720 ($\pm$ 0.012) & 0.836 ($\pm$ 0.008) & 0.809 ($\pm$ 0.021) & \textbf{0.868} ($\pm$ 0.012) \\
& Tooth & \textbf{0.865} ($\pm$ 0.005) & \textbf{0.927} ($\pm$ 0.003) & 0.928 ($\pm$ 0.004) & \textbf{0.927} ($\pm$ 0.008) & 0.856 ($\pm$ 0.006) & 0.922 ($\pm$ 0.004) & \textbf{0.930} ($\pm$ 0.003) & 0.915 ($\pm$ 0.005) \\
& Pulp & \textbf{0.598} ($\pm$ 0.004) & \textbf{0.747} ($\pm$ 0.003) & \textbf{0.752} ($\pm$ 0.009) & 0.748 ($\pm$ 0.008) & 0.586 ($\pm$ 0.010) & 0.738 ($\pm$ 0.008) & 0.696 ($\pm$ 0.015) & \textbf{0.790} ($\pm$ 0.010) \\
& Dentin & 0.784 ($\pm$ 0.008) & 0.879 ($\pm$ 0.005) & 0.917 ($\pm$ 0.004) & 0.844 ($\pm$ 0.007) & \textbf{0.857} ($\pm$ 0.007) & \textbf{0.923} ($\pm$ 0.004) & \textbf{0.937} ($\pm$ 0.002) & \textbf{0.910} ($\pm$ 0.010) \\
& Enamel & 0.739 ($\pm$ 0.013) & 0.844 ($\pm$ 0.010) & \textbf{0.841} ($\pm$ 0.008) & 0.854 ($\pm$ 0.018) & \textbf{0.749} ($\pm$ 0.009) & \textbf{0.850} ($\pm$ 0.006) & 0.831 ($\pm$ 0.017) & \textbf{0.875} ($\pm$ 0.020) \\
& Composite & \textbf{0.416} ($\pm$ 0.018) & \textbf{0.512} ($\pm$ 0.019) & \textbf{0.610} ($\pm$ 0.014) & 0.458 ($\pm$ 0.024) & 0.414 ($\pm$ 0.022) & 0.511 ($\pm$ 0.017) & 0.540 ($\pm$ 0.043) & \textbf{0.513} ($\pm$ 0.043) \\
\midrule
\multirow{9}{*}{\textbf{HRNet}} & Average & 0.671 ($\pm$ 0.006) & 0.779 ($\pm$ 0.005) & \textbf{0.787} ($\pm$ 0.003) & 0.785 ($\pm$ 0.007) & \textbf{0.695} ($\pm$ 0.003) & \textbf{0.795} ($\pm$ 0.002) & 0.782 ($\pm$ 0.002) & \textbf{0.821} ($\pm$ 0.002) \\
& Background & 0.978 ($\pm$ 0.001) & 0.989 ($\pm$ 0.000) & 0.991 ($\pm$ 0.001) & 0.987 ($\pm$ 0.001) & \textbf{0.979} ($\pm$ 0.000) & \textbf{0.989} ($\pm$ 0.000) & \textbf{0.991} ($\pm$ 0.001) & \textbf{0.987} ($\pm$ 0.001) \\
& Upper & 0.514 ($\pm$ 0.006) & 0.677 ($\pm$ 0.005) & 0.625 ($\pm$ 0.011) & 0.753 ($\pm$ 0.017) & \textbf{0.526} ($\pm$ 0.004) & \textbf{0.686} ($\pm$ 0.003) & \textbf{0.625} ($\pm$ 0.009) & \textbf{0.772} ($\pm$ 0.013) \\
& Lower & 0.704 ($\pm$ 0.012) & 0.826 ($\pm$ 0.008) & \textbf{0.818} ($\pm$ 0.012) & 0.837 ($\pm$ 0.020) & \textbf{0.717} ($\pm$ 0.004) & \textbf{0.835} ($\pm$ 0.002) & 0.814 ($\pm$ 0.011) & \textbf{0.860} ($\pm$ 0.007) \\
& Tooth & 0.844 ($\pm$ 0.006) & 0.915 ($\pm$ 0.003) & 0.916 ($\pm$ 0.002) & 0.915 ($\pm$ 0.005) & \textbf{0.852} ($\pm$ 0.003) & \textbf{0.920} ($\pm$ 0.001) & \textbf{0.924} ($\pm$ 0.004) & \textbf{0.916} ($\pm$ 0.002) \\
& Pulp & 0.501 ($\pm$ 0.014) & 0.666 ($\pm$ 0.012) & \textbf{0.643} ($\pm$ 0.003) & 0.697 ($\pm$ 0.026) & \textbf{0.515} ($\pm$ 0.007) & \textbf{0.679} ($\pm$ 0.006) & 0.612 ($\pm$ 0.011) & \textbf{0.769} ($\pm$ 0.008) \\
& Dentin & 0.731 ($\pm$ 0.007) & 0.844 ($\pm$ 0.005) & 0.901 ($\pm$ 0.008) & 0.795 ($\pm$ 0.002) & \textbf{0.822} ($\pm$ 0.004) & \textbf{0.902} ($\pm$ 0.003) & \textbf{0.929} ($\pm$ 0.001) & \textbf{0.877} ($\pm$ 0.005) \\
& Enamel & 0.687 ($\pm$ 0.012) & 0.809 ($\pm$ 0.008) & \textbf{0.809} ($\pm$ 0.009) & 0.817 ($\pm$ 0.025) & \textbf{0.729} ($\pm$ 0.002) & \textbf{0.838} ($\pm$ 0.002) & 0.807 ($\pm$ 0.011) & \textbf{0.876} ($\pm$ 0.016) \\
& Composite & 0.410 ($\pm$ 0.015) & 0.509 ($\pm$ 0.015) & \textbf{0.591} ($\pm$ 0.017) & 0.476 ($\pm$ 0.036) & \textbf{0.418} ($\pm$ 0.019) & \textbf{0.515} ($\pm$ 0.023) & 0.553 ($\pm$ 0.019) & \textbf{0.512} ($\pm$ 0.041) \\
\bottomrule
\end{tabular}
}
\label{tab:test_results}
\end{table}

\end{document}